\documentclass[runningheads]{llncs}

\usepackage{eccv}

\usepackage{eccvabbrv}

\usepackage{graphicx}
\usepackage{booktabs}

\usepackage{amsmath}
\usepackage{paracol}
\usepackage{multirow}
\usepackage{amssymb}
\usepackage{subcaption}
\usepackage{float}
\usepackage{placeins}
\usepackage[export]{adjustbox}
\usepackage{comment}

\usepackage[accsupp]{axessibility}  

\usepackage{hyperref}

\usepackage{orcidlink}

\hypersetup{
    pdftitle={A Systematic Comparison of Training Objectives for Out-of-Distribution Detection in Image Classification},
    pdfauthor={Furkan Genç, Onat Özdemir, Emre Akbaş}
}

\begin{document}

\title{A Systematic Comparison of Training Objectives for Out-of-Distribution Detection in Image Classification}
\titlerunning{A Systematic Comparison of Training Objectives for OOD Detection}

\author{Furkan Genç\inst{1}\orcidlink{0009-0005-8651-7194} \and
Onat Özdemir\inst{2}\orcidlink{0009-0005-8527-6253} \and
Emre Akbaş\inst{2}\orcidlink{0000-0002-3760-6722}}

\authorrunning{Genç et al.}

\institute{Dept. of Electrical and Electronics Engineering, Bilkent University \email{furkan.genc@bilkent.edu.tr} \and
Dept. of Computer Engineering, Middle East Technical University
}

\maketitle
\begin{abstract}
Out-of-distribution (OOD) detection is critical in safety-sensitive applications. While this challenge has been addressed from various perspectives, the influence of training objectives on OOD behavior remains comparatively underexplored. In this paper, we present a systematic comparison of four widely used training objectives: Cross-Entropy Loss, Prototype Loss, Triplet Loss, and Average Precision (AP) Loss, spanning probabilistic, prototype-based, metric-learning, and ranking-based supervision, for OOD detection in image classification under standardized OpenOOD protocols. Within the evaluated ResNet-18/OpenOOD setting and objective-specific OOD scoring rules, Cross-Entropy Loss, Prototype Loss, and AP Loss achieve comparable in-distribution accuracy, while Cross-Entropy Loss provides the most consistent near- and far-OOD AUROC overall; the other objectives can be competitive in specific settings.
\end{abstract}

\section{Introduction}
Machine learning models have achieved strong performance in domains such as computer vision, natural language processing, and speech recognition. However, real-world deployments often encounter inputs that differ from the training distribution, making \textbf{out-of-distribution (OOD)} detection essential for reliability and safety. This is especially important in safety-critical applications such as autonomous driving, medical diagnosis, and security systems, where OOD inputs can cause unpredictable model behavior.

Most OOD detection work focuses on post-processing scores \cite{hendrycks2017baseline, lee2018simple, liang2018enhancing, liu2020energy, yang2024generalized} or specialized training strategies \cite{hendrycks2019oe, hendrycks2020augmix}. Although some studies analyze representation or pretraining objectives, such as contrastive learning and masked image modeling, systematic comparisons among common supervised objectives remain limited \cite{sun2022ood_knn,li2023rethinking,tack2020csi,yang2024generalized}.

This paper systematically compares supervised training objectives for OOD detection in image classification under the standardized OpenOOD evaluation framework \cite{zhang2024openood}. We take Cross-Entropy Loss \cite{goodfellow2016deep} as the standard classification objective and compare it with three common alternatives: Triplet Loss \cite{schroff2015facenet}, Prototype Loss \cite{snell2017prototypical}, and Average Precision (AP) Loss \cite{ap_loss_cvpr19,oksuz2021rank}. Although these objectives are successful in metric learning, few-shot learning, and ranking-based training \cite{schroff2015facenet,hermans2017defense,snell2017prototypical,ap_loss_cvpr19,brown2020smooth}, their relative OOD behavior is less systematically characterized than representation/pretraining objectives or OOD-specific losses \cite{sun2022ood_knn,li2023rethinking,ming2023hyperspherical,lu2024mixture,yang2024generalized}.

Triplet Loss, introduced in FaceNet \cite{schroff2015facenet}, compares an anchor input to a positive input (same class) and a negative input (different class). By minimizing the distance between the anchor and the positive and maximizing the distance to the negative, the model learns to recognize similarities and differences in the embedding space. Prototype Loss, proposed by Snell et al. \cite{snell2017prototypical}, represents each class by a prototype embedding and trains the model to minimize the distance between a sample’s embedding and its corresponding class prototype, while maximizing separation from the prototypes of other classes. Average Precision (AP) Loss \cite{ap_loss_cvpr19,oksuz2021rank} is a ranking-based objective that optimizes the ordering of scores by promoting higher scores for positive (in-class) examples than for negatives, aiming to improve separability in terms of score rankings.

Most OOD baselines train classifiers with Cross-Entropy Loss and then study post-hoc confidence scores \cite{hendrycks2017baseline,liang2018enhancing,lee2018simple,liu2020energy}. Standardized benchmarks such as OpenOOD \cite{yang2022openood} and OpenOOD v1.5 \cite{zhang2024openood} enable fairer evaluation across near- and far-OOD shifts, yet the role of the training objective within these protocols remains underexplored.

We evaluate these objectives under the OpenOOD benchmark to compare their ID accuracy and OOD detection performance under the corresponding inference rules. The selected losses represent probabilistic classification, metric learning, prototype-based learning, and ranking-based optimization, each paired with a standard inference rule. This enables a controlled comparison under a fixed backbone and standardized protocols, while avoiding objectives that require additional components such as specialized augmentations, large-batch sampling, or extra regularization.

The main contributions of this paper are:

\begin{itemize}
    \item We provide a systematic comparison of four widely used training objectives: Cross-Entropy, Triplet, Prototype, and Average Precision losses, covering probabilistic, metric-learning, prototype-based, and ranking-based supervision paradigms for image classification.
    \item We use the OpenOOD benchmark \cite{zhang2024openood} with a fixed backbone and standardized protocols to control the architecture, datasets, and evaluation procedure, enabling fair and reproducible comparisons across objectives.
    \item We analyze the trade-offs observed across the evaluated objective-inference pairings in terms of in-distribution accuracy and near- and far-OOD detection, offering practical guidance within the studied setting.
\end{itemize}

Within the studied setting, specialized objectives can be competitive in specific cases, but Cross-Entropy remains the most reliable overall default for ID accuracy and OOD detection.

\section{Related Work}
Out-of-distribution (OOD) detection is a critical area of machine learning research, focusing on models' ability to recognize and appropriately handle inputs that differ significantly from the training data. This section reviews key studies on OOD detection, emphasizing training objectives, evaluation benchmarks like OpenOOD, and methodologies relevant to our work.

\subsection{Out-of-Distribution Detection Methods}

Early work \cite{hendrycks2017baseline} introduced \emph{maximum softmax probability} as a simple OOD baseline, showing that existing classifiers can provide useful detection scores. ODIN \cite{liang2018enhancing} improves softmax-based detection with temperature scaling and input perturbations, while Lee et al. \cite{lee2018simple} use Mahalanobis confidence scores computed from layer activations. Other post-hoc scores include energy-based scoring \cite{liu2020energy}, the strengthened softmax-based GEN score \cite{liu2023gen}, and non-parametric nearest-neighbor distances in deep feature space, which can be effective when simple parametric assumptions fail \cite{sun2022ood_knn}. Complementary training-time strategies include Outlier Exposure \cite{hendrycks2019oe}, which uses auxiliary outlier data, and AugMix \cite{hendrycks2020augmix}, which improves robustness through augmentation.

\subsection{Training Objectives}

Cross-Entropy Loss remains the standard loss function for classification tasks due to its effectiveness in optimizing probabilistic models \cite{goodfellow2016deep}. However, it does not explicitly enforce intra-class compactness or large inter-class separation, which may limit its effectiveness in OOD detection when ID and OOD classes are not well separated.

Confidence-aware variants of Cross-Entropy have also been studied. Focal Loss reweights the Cross-Entropy objective to emphasize harder examples \cite{lin2017focal}, and subsequent work has shown that it can also affect classifier calibration \cite{mukhoti2020calibrating,kimura2025geometric}. We consider such variants part of the broader probabilistic-classification family and leave comparisons among objectives within that family outside the scope of our cross-paradigm study.

Metric learning losses, such as Contrastive Loss and Triplet Loss, learn embeddings where similar instances are closer and dissimilar ones are farther apart \cite{schroff2015facenet, chopra2005learning}. Triplet Loss has been successful in face recognition and person re-identification \cite{hermans2017defense}, but is less commonly studied for OOD detection.

Prototypical Networks \cite{snell2017prototypical} and related prototype-based methods have been widely used in few-shot learning \cite{ravichandran2019few,zhou2023revisiting}. Learning a prototype for each class can generalize these methods to new classes with limited samples.

Average Precision Loss \cite{ap_loss_cvpr19,oksuz2021rank}, which directly optimizes ranking metrics such as the average precision, has been used in object detection \cite{ap_loss_cvpr19, oksuz2021rank}, image segmentation \cite{oksuz2021rank}, image retrieval \cite{brown2020smooth}, edge detection \cite{cetinkaya2024ranked} and imbalanced image classification tasks \cite{baran_gulmez_thesis}. Although Average Precision Loss can be adapted for OOD detection to improve ID-OOD ranking, it remains less commonly adopted than standard classification-based objectives.

Prior work has also investigated how training or pre-training objectives influence OOD behavior. Sun et al.~\cite{sun2022ood_knn} study nearest-neighbor OOD detection and report differences between contrastive-style and CE representations, while Li et al.~\cite{li2023rethinking} analyze masked image modeling pre-training. Other works design OOD-specific objectives that shape embedding geometry, such as hyperspherical embeddings~\cite{ming2023hyperspherical} or mixtures of prototypes~\cite{lu2024mixture}. Our work complements these efforts by comparing four common supervised objectives under a fixed architecture and standardized OpenOOD evaluation.

\subsection{OpenOOD Benchmark}

Standardized benchmarks such as OpenOOD \cite{yang2022openood} have advanced OOD evaluation by defining common datasets, protocols, and metrics. OpenOOD v1.5 \cite{zhang2024openood} extends this framework with additional datasets and evaluation settings, enabling fairer comparisons across methods.

OpenOOD categorizes OOD datasets into \textbf{near-OOD} and \textbf{far-OOD} splits based on similarity to the ID data. Near-OOD samples are semantically or visually closer to ID classes, whereas far-OOD samples differ more substantially from the ID distribution. We use this benchmark to compare the evaluated objectives under standardized distribution shifts.

\section{Methodology}
This section describes our methodology, encompassing the model architectures, training objectives, out-of-distribution (OOD) detection strategies, and inference procedures.

\subsection{Model Architecture}
We employ ResNet-18 \cite{he2016deep} using the default OpenOOD implementations \cite{zhang2024openood}. 
For objectives that operate on class logits (Cross-Entropy and AP), we set the final fully connected (FC) layer to output $C$ logits, where $C$ is the number of in-distribution classes. 
For embedding-based objectives (Triplet and Prototype), we replace the final FC layer with an embedding head of dimension $\mathrm{ED}$ and treat its output as $f(x)$ for metric/prototype inference. 
For CIFAR-10/100, the initial layers are modified to accommodate $32\times 32$ images, whereas for ImageNet-200, we use the standard ResNet-18 architecture.

\subsection{Training Objectives and Inference}
\label{sec:training_objectives_inference}
In our experiments, we compare four training objectives: Cross-Entropy Loss \cite{goodfellow2016deep}, Triplet Loss \cite{schroff2015facenet}, Prototype Loss \cite{snell2017prototypical}, and Average Precision (AP) Loss \cite{ap_loss_cvpr19,oksuz2021rank}.
The selected hyperparameters and the OOD post-processing rule used for each dataset-objective pair are summarized in Appendix~\ref{app:selected_params}.

We chose these objectives to represent common supervision paradigms: probabilistic alignment (Cross-Entropy), margin-based metric learning (Triplet), prototype clustering (Prototype), and ranking-based supervision (AP).
Cross-Entropy trains a classifier by maximizing the likelihood of the ground-truth class under a softmax distribution over logits. Prototype training replaces logits with distances to learned class prototypes, encouraging intra-class compactness around each prototype. Triplet training learns an embedding space by enforcing a margin between anchor-positive and anchor-negative distances. Finally, AP training uses ranking-based supervision to promote higher scores for positives than negatives, aiming to improve ordering quality.

We define all OOD scores so that higher values indicate greater OOD likelihood. Because these objectives produce different natural outputs (e.g., logits vs.\ embeddings), we evaluate each model using OOD scoring rule(s) that are standard and naturally aligned with the corresponding objective family. For Cross-Entropy and Prototype training, we consider MSP-based scoring (i.e., $-\max_{c} p_c(x)$) and predictive-entropy scoring and report the selected rule for each dataset-objective pair in Appendix~\ref{app:selected_params}. For Triplet training, we use the standard nearest-neighbor distance in the learned embedding space (i.e., the minimum distance to the training embeddings). For AP training, we treat the per-class scores $s_c(x)$ as logits, compute class probabilities via softmax, and use predictive entropy (Eq.~\ref{eq:entropy}).

Accordingly, our comparison is intentionally scoped to \emph{objective-inference pairings} that are standard in practice, including probability-based scoring (such as MSP/entropy for CE/Prototype/AP) and distance-to-training-embeddings for Triplet. This design mirrors real deployment, where the training objective dictates the natural form of confidence used for OOD scoring. While a single universal detector could further isolate the objective effect, it would require additional modeling assumptions (e.g., converting embeddings to calibrated class probabilities or fitting density models), confounding the study with design choices beyond the loss itself. Consequently, the reported OOD results should be interpreted as comparisons of objective-inference pairings rather than as fully factorized estimates of the training-objective effect.

\subsubsection{Cross-Entropy Loss}

Cross-Entropy Loss is a widely used loss function for multi-class classification tasks. It measures the discrepancy between the predicted probability distribution and the actual distribution. The loss for a single sample is defined as:
\begin{equation}
    L_{\text{CE}} = -\sum_{c=1}^{C} y_c \log(p_c)
\end{equation}
where $C$ is the number of classes, $y$ is the ground truth label encoded as a one-hot vector, and $y_c$ is its $c^\mathrm{th}$ element, i.e. we have $y=[y_1, y_2, \dots, y_C]$. 
$p_c$ is the predicted probability for class $c$, obtained by applying the softmax function to the network's output logits.

During inference, the class with the highest probability is selected as the predicted label: 
\begin{equation}
    \hat{y}(x) = \arg\max_{c} p_c(x)
\end{equation}
We test two OOD scoring methods: (i) Maximum Softmax Probability (MSP) \cite{hendrycks2017baseline}:
\begin{equation}
    S_{\text{OOD}}(x) = -\max_{c} p_c(x)
\label{eq:msp}
\end{equation}
 
\noindent and (ii) Entropy of Softmax Probabilities:
\begin{equation}
    S_{\text{OOD}}(x) = -\sum_{c=1}^{C} p_c(x) \log p_c(x).
\label{eq:entropy}
\end{equation}

\subsubsection{Triplet Loss}
\label{sec:triplet_loss}

Triplet Loss \cite{schroff2015facenet} is a metric learning loss function designed to learn an embedding space where samples of the same class are closer together and samples of different classes are farther apart. The loss for a triplet of samples (anchor $a$, positive $p$, negative $n$) is defined as:

\begin{equation}
    L_{\text{Triplet}} = \max\left(0, D(f(a), f(p)) - D(f(a), f(n)) + \alpha\right)
\end{equation}

\noindent where $f(\cdot)$ is the learned embedding function, $D(u, v)$ is the squared Euclidean distance, and $\alpha$ is the margin hyperparameter ensuring minimum class separation in the embedding space. In all triplet-loss experiments, we use a fixed margin of $\alpha = 1$.

The Triplet Loss objective \cite{schroff2015facenet} encourages triplets to satisfy
\begin{equation}
\|f(x_a)-f(x_p)\|_2^2 + \alpha < \|f(x_a)-f(x_n)\|_2^2.
\end{equation}
We utilize a Random Negative Triplet Selector for CIFAR-10 \cite{krizhevsky2009learning}, which employs random triplet sampling. However, for CIFAR-100 \cite{krizhevsky2009learning} and ImageNet-200 \cite{van2020scan, deng2009imagenet}, we employ a Semi-Hard Negative Triplet Selector. In this approach, negatives are chosen so that they are farther from the anchor than the positive sample, but still within the margin.
Let $d_{ap}=\|f(x_a)-f(x_p)\|_2^2$ and $d_{an}=\|f(x_a)-f(x_n)\|_2^2$. Semi-hard negatives satisfy
\begin{equation}
    d_{ap} < d_{an} < d_{ap} + \alpha.
\end{equation}
The mining strategy was selected using ID validation accuracy. Random mining was preferred for CIFAR-10, whereas semi-hard mining was preferred for CIFAR-100 and ImageNet-200, where more targeted negative selection becomes useful as class diversity increases.

The predicted class is determined by identifying the nearest training embedding:
\begin{equation}
    \hat{y}(x) = y\left(\arg\min_{x_i^\text{train}} D(f(x), f(x_i^{\text{train}}))\right),
\end{equation}
where $y(\cdot)$ returns the ground-truth label of the example. OOD Score is computed as the minimum distance to the training embeddings:
\begin{equation}
    S_{\text{OOD}}(x) = \min_{i} D(f(x), f(x_i^{\text{train}})).
\end{equation}
This distance-to-training-set score is closely related to deep nearest-neighbor OOD detection methods \cite{sun2022ood_knn}.

\subsubsection{Prototype Loss}

For the prototype learning approach, we employ the \emph{Generalized Convolutional Prototype Learning (GCPL)} method \cite{yang2018robust}. 
The prototype loss in GCPL has two components with a balancing hyperparameter $\lambda$:

\begin{equation}
L_{\text{Total}} = L_{\text{DCE}} + \lambda L_{\text{Center}},
\end{equation}

\noindent where
$L_{\text{DCE}}$ is the distance-based cross-entropy loss, and $L_{\text{Center}}$ is the center loss promoting intra-class compactness.
$L_{\text{DCE}}$ is defined as 
\begin{equation}
L_{\text{DCE}} = -\sum_{i=1}^{N} \log \frac{\exp\left(-\tau D(f(x_i), m_{y_i})\right)}{\sum_{k=1}^{C} \exp\left(-\tau D(f(x_i), m_{k})\right)}
\end{equation}
where $f(x_i)$ is the embedding of example $x_i$, 
$m_{k}$ is the prototype (mean embedding) of class $k$,  
$D(\cdot,\cdot)$ is the squared Euclidean distance. 
$\tau$ controls the sharpness of the probability distribution and is selected using ID validation accuracy, with the selected values reported in Appendix~\ref{app:selected_params}.
The center loss is defined as:
\begin{equation}
L_{\text{Center}} = \frac{1}{N} \sum_{i=1}^{N} \left\| f(x_i) - m_{y_i} \right\|_2^2.
\end{equation}
The predicted class is determined by the nearest class prototype:
\begin{equation}
\hat{y}(x) = \arg\min_{c} D(f(x), m_{c})
\end{equation}
We test two OOD scoring methods as in the Cross-Entropy setting: (i) MSP as in Equation~\eqref{eq:msp} and (ii) entropy as in Equation~\eqref{eq:entropy}. In both cases, the class probability $p_c(x)$ is computed using the selected temperature parameter $\tau$:

\begin{equation}
p_c(x) = \frac{\exp(-\tau D(f(x), m_{c}))}{\sum_{k=1}^{C} \exp(-\tau D(f(x), m_{k}))}.
\end{equation}

\subsubsection{Average Precision (AP) Loss }

The Average Precision Loss \cite{ap_loss_cvpr19,oksuz2021rank} is a ranking-based loss, which is designed to directly optimize the Average Precision (AP) evaluation measure. For binary classification, the AP Loss is defined as 1 minus average precision: 

\begin{equation}
L_{\text{AP}} = 1 - \frac{1}{|P|} \sum_{i \in P} \text{prec}(i), 
\end{equation}

\noindent where prec(i) is the precision of the $i^\text{th}$ positive example. This can be expressed as the ratio of the rank of the $i^\text{th}$ positive example among all positives and the rank of the same example among all examples (negatives included):

\begin{equation}
    \text{prec}(i) = \frac{\text{rank}_{\text{pos}}(i)}{\text{rank}(i)}. 
\end{equation}

The rank($\cdot$) operator induces discrete ordering via pairwise comparisons and is not differentiable in its exact form. Following \cite{oksuz2021rank}, we implement AP-style optimization using the standard three-step ranking-loss formulation over score differences and incorporate the paper's identity/error-driven update for the non-differentiable primary-term step within backpropagation, which provides well-defined gradients with respect to logits and enables end-to-end optimization with SGD.

We adopt a one-vs-all approach for each class $c$ to extend the AP Loss to multi-class classification. For each class, we treat samples belonging to class $c$ as positives ($y_i = 1$) and all other samples as negatives ($y_i = 0$). We compute the AP Loss for each class and then sum over all classes:
\begin{equation}
    L_{\text{Total}} = \sum_{c=1}^{C} L_{\text{AP}}^{(c)}
\end{equation}

\noindent For other details, the reader is referred to \cite{ap_loss_cvpr19, oksuz2021rank}. The predicted class is the arg-max of $p_c(x)$ where 
\begin{equation}
    p_c(x) = \frac{\exp(s_c(x))}{\sum_{k=1}^{C} \exp(s_k(x))}. 
\end{equation}

\noindent OOD score is the same as Equation \eqref{eq:entropy}.

\section{Experiments}
In this section, we evaluate four training objectives: Cross-Entropy Loss, Triplet Loss, Prototype Loss, and Average Precision (AP) Loss on standard image classification datasets for Out-of-Distribution (OOD) detection.

\paragraph{Datasets.}
We train models on three datasets: CIFAR-10 \cite{krizhevsky2009learning}, CIFAR-100 \cite{krizhevsky2009learning} and ImageNet-200 \cite{van2020scan, deng2009imagenet}. For these three datasets, which are also referred to as the ID (in-distribution) dataset, the OpenOOD benchmark defines the following near- and far-OOD datasets: For \textbf{CIFAR-10}, near-OOD: CIFAR-100, TinyImageNet \cite{tinyimagenet}; far-OOD: MNIST \cite{deng2012mnist}, SVHN \cite{netzer2011reading}, Texture \cite{kylberg2011kylberg}, Places365 \cite{zhou2017places}. 
For \textbf{CIFAR-100}, near-OOD: CIFAR-10, TinyImageNet; far-OOD: MNIST, SVHN, Texture, Places365.
For \textbf{ImageNet-200} \cite{deng2009imagenet}, near-OOD: SSB-hard \cite{vaze2022openset}, NINCO \cite{bitterwolf2023ninco}; far-OOD: iNaturalist \cite{van2018inaturalist}, Texture, OpenImage-O \cite{wang2022vim}.

\paragraph{Evaluation Metrics.}
To evaluate models, we use the following metrics from the OpenOOD benchmark: 
In-Distribution (ID) accuracy is the classification accuracy on the test set of the ID dataset. 
Area Under the Receiver Operating Characteristic Curve (AUROC) measures the model's ability to distinguish between ID and OOD samples across all classification thresholds. We report both near-OOD AUROC and far-OOD AUROC. A higher AUROC indicates better discrimination between ID and OOD samples. We focus on AUROC as a threshold-free ranking metric; operating-point and precision-recall metrics such as FPR95 and AUPR are not evaluated in this study.

\paragraph{Implementation Details.}
All experiments are implemented using PyTorch \cite{paszke2019pytorch}. We use the OpenOOD framework \cite{zhang2024openood} and its predefined train, validation, and test splits to standardize dataset loading and evaluation. Model and training hyperparameters are selected using ID validation accuracy. When multiple OOD scoring rules are considered for Cross-Entropy and Prototype training, MSP versus predictive entropy is selected using AUROC on the dedicated OpenOOD ID-validation and OOD-validation splits. The reported near- and far-OOD test splits are used only for final evaluation and do not influence hyperparameter or scoring-rule selection. The selected hyperparameters and OOD scoring rules are reported in Appendix~\ref{app:selected_params}.

\paragraph{Training Configuration.}
We use stochastic gradient descent (SGD) as an optimizer with a momentum of 0.9 and a weight decay of \(5 \times 10^{-4}\). We set the batch size to 128 for CIFAR-10 and CIFAR-100, and 256 for ImageNet-200. We use a cosine annealing learning rate schedule without warm restarts \cite{loshchilov2016sgdr}, with an initial learning rate chosen for each loss function. We train the models for 100 epochs on CIFAR-10 and CIFAR-100, and for 90 epochs on ImageNet-200, following the default OpenOOD training schedule for this benchmark to keep our setup aligned with the standard evaluation protocol. For each loss function, we perform hyperparameter tuning over a range of learning rates and other loss-specific parameters.

\section{Results}
We run five independent experiments for each objective and dataset, and present bar charts sorted from highest to lowest performance for each metric. Adjacent methods in the sorted order are compared using Welch's $t$-tests ($p<0.05$) over the five-run distributions, with significant differences marked by (**).

We use Welch's $t$-test rather than Student's $t$-test since it does not assume equal variances and is appropriate for comparing mean performance across independent training runs where variability can differ across objectives \cite{welch1947general}. 
We do not apply McNemar's test \cite{mcnemar1947note} as it is designed for paired comparisons of discrete decisions on the same examples at a fixed operating point, whereas our primary metrics are computed from continuous confidence scores (notably AUROC) and are summarized at the run level.

We show three plots per dataset: (a) ID Accuracy, (b) near-OOD AUROC, and (c) far-OOD AUROC. The abbreviations are AP for AP Loss, CE for Cross-Entropy Loss, PT for Prototype Loss, and TL for Triplet Loss. Unless otherwise stated, we report results using the selected post-processing rule for each dataset-objective pair listed in Appendix~\ref{app:selected_params}; complete numeric results are provided in Appendix~\ref{app:extended_quant}. For brevity, objective names in the Results and Discussion sections refer to the corresponding objective-inference pairings defined in Section~\ref{sec:training_objectives_inference}.

\subsection{CIFAR-10 Results}
\begin{figure}[!t]
    \centering
    \hfill
    \begin{subfigure}[t]{0.21\linewidth}
        \centering
        \includegraphics[width=\linewidth]{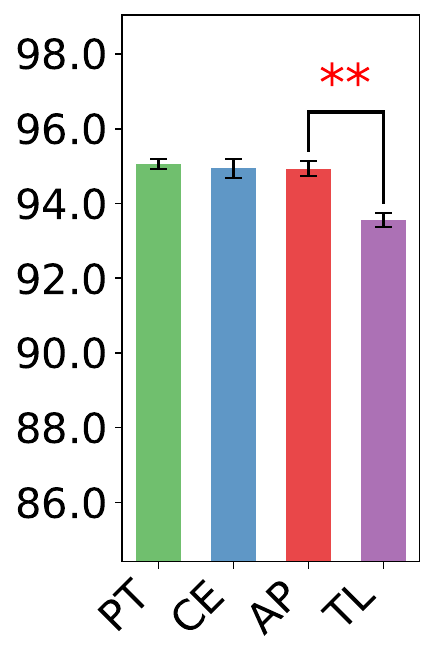}
        \caption{}            
    \end{subfigure}
    \hfill 
    \begin{subfigure}[t]{0.21\linewidth}
        \centering
        \includegraphics[width=\linewidth]{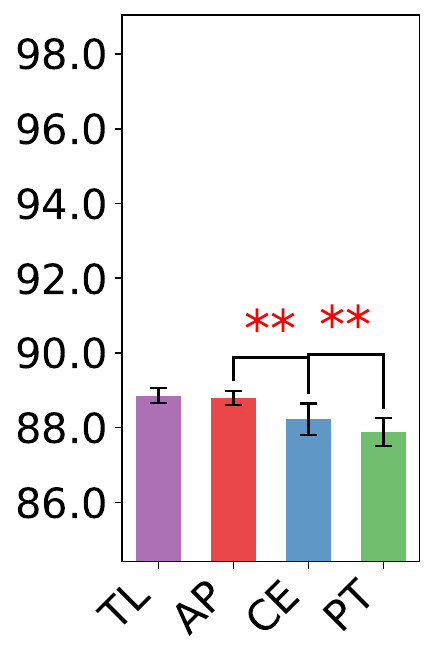}
        \caption{} 
    \end{subfigure}
    \hfill 
    \begin{subfigure}[t]{0.21\linewidth}
        \centering
        \includegraphics[width=\linewidth]{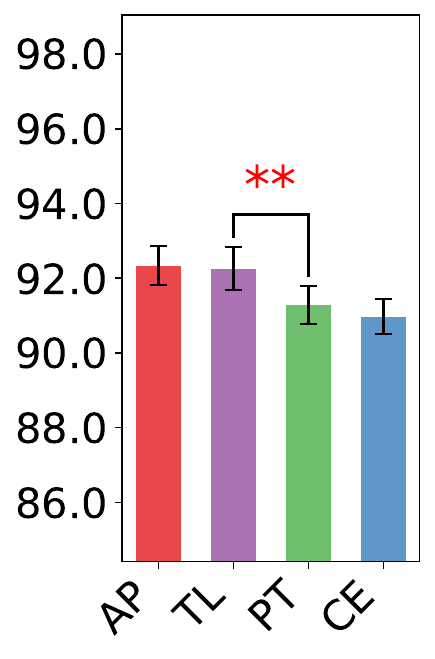}
        \caption{} 
    \end{subfigure}
    \hfill
    \vspace{-0.5em}
    \caption{Results for \textbf{CIFAR-10}: (a) ID Accuracy, (b) near-OOD AUROC, and (c) far-OOD AUROC. ** denotes statistically significant difference.}
    \label{fig:cifar10_results}
\end{figure}

As shown in Figure~\ref{fig:cifar10_results}, on CIFAR-10, 
Prototype, Cross-Entropy, and AP achieve similar top ID accuracies (95.06\%, 94.95\%, and 94.93\%), with no significant differences among them.  
AP and Triplet provide the strongest OOD detection overall: Triplet achieves the best near-OOD AUROC (88.85\%), while AP achieves the best far-OOD AUROC (92.34\%), with each being second-best on the other OOD setting (AP: 88.80\% near; Triplet: 92.25\% far).
While Triplet maintains competitive OOD detection performance (88.85\% near, 92.25\% far), it performs the worst in ID accuracy (93.56\%). 
Overall, none of the training objectives consistently dominates across all three metrics.

\FloatBarrier

\subsection{CIFAR-100 Results}
\begin{figure}[!t]
    \centering
    \hfill
    \begin{subfigure}[t]{0.21\linewidth}
        \centering
        \includegraphics[width=\linewidth]{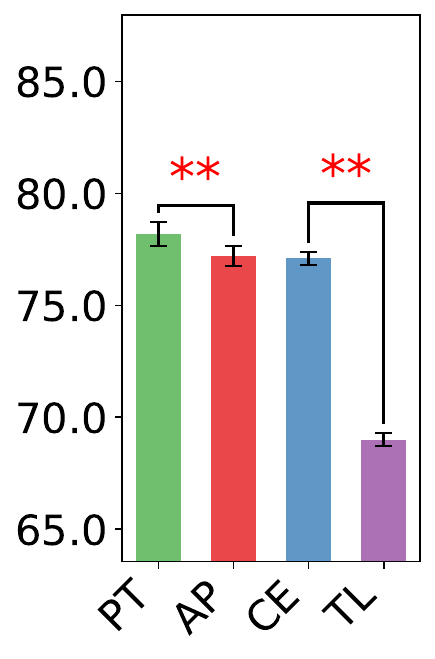}
        \caption{}            
    \end{subfigure}
    \hfill 
    \begin{subfigure}[t]{0.21\linewidth}
        \centering
        \includegraphics[width=\linewidth]{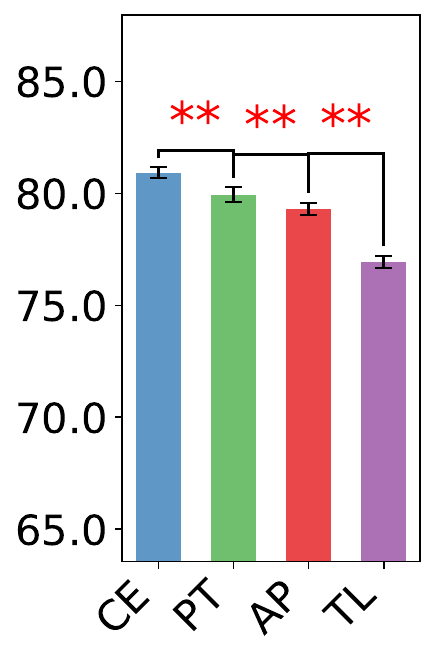}
        \caption{} 
    \end{subfigure}
    \hfill 
    \begin{subfigure}[t]{0.21\linewidth}
        \centering
        \includegraphics[width=\linewidth]{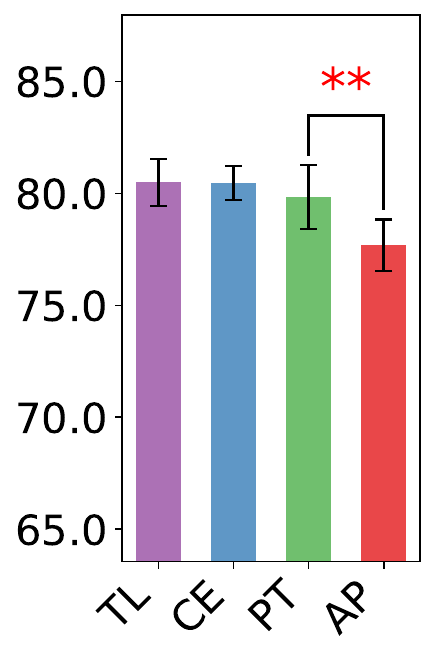}
        \caption{} 
    \end{subfigure}
    \hfill
    \vspace{-0.5em}
    \caption{Results for \textbf{CIFAR-100}: (a) ID Accuracy, (b) near-OOD AUROC, and (c) far-OOD AUROC. ** denotes statistically significant difference.}
    \label{fig:cifar100_results}
\end{figure}

Referring to Figure~\ref{fig:cifar100_results}, on CIFAR-100, Cross-Entropy achieves the best near-OOD AUROC (80.94\%) and competitive far-OOD AUROC (80.46\%), giving the strongest overall balance as class count increases. Prototype achieves the highest ID accuracy (78.18\%), indicating effective intra-class compactness and discriminative feature learning. AP shows moderate near-OOD detection (79.32\%) but weaker far-OOD detection (77.68\%) and slightly lower ID accuracy (77.21\%). Triplet attains the best far-OOD AUROC (80.50\%) but has substantially lower ID accuracy (69.00\%) and weaker near-OOD detection (76.95\%), showing that this configuration can be competitive for some far-OOD shifts while performing substantially worse on ID accuracy and near-OOD detection.

\FloatBarrier

\subsection{ImageNet-200 Results}
\begin{figure}[!t]
    \centering
    \hfill
    \begin{subfigure}[t]{0.21\linewidth}
        \centering
        \includegraphics[width=\linewidth]{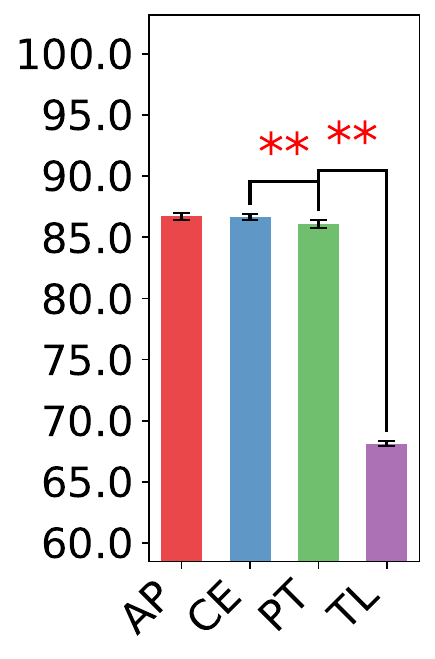}
        \caption{}            
    \end{subfigure}
    \hfill 
    \begin{subfigure}[t]{0.21\linewidth}
        \centering
        \includegraphics[width=\linewidth]{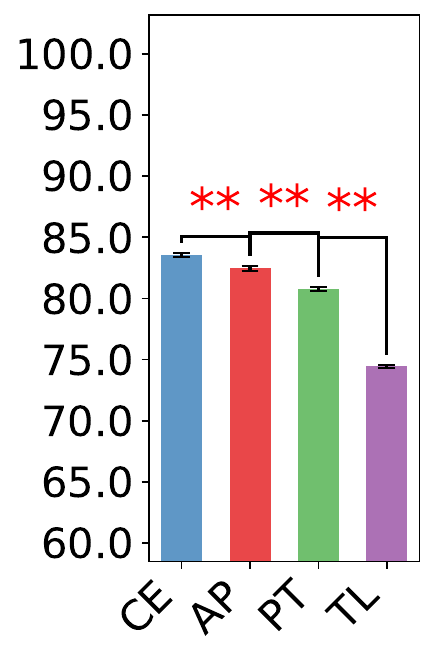}
        \caption{} 
    \end{subfigure}
    \hfill 
    \begin{subfigure}[t]{0.21\linewidth}
        \centering
        \includegraphics[width=\linewidth]{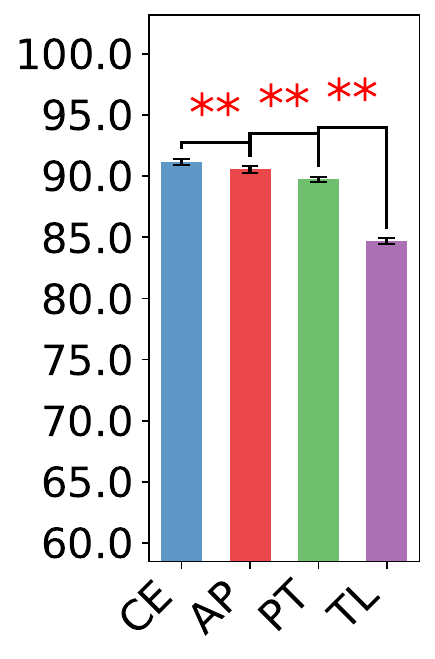}
        \caption{} 
    \end{subfigure}
    \hfill
    \vspace{-0.5em}
    \caption{Results for \textbf{ImageNet-200}: (a) ID Accuracy, (b) near-OOD AUROC, and (c) far-OOD AUROC. ** denotes statistically significant difference.}
    \label{fig:imagenet200_results}
\end{figure}

Figure~\ref{fig:imagenet200_results} shows that on ImageNet-200, Cross-Entropy gives the strongest OOD detection (83.54\% near, 91.16\% far) while nearly matching AP in ID accuracy (86.66\% vs. 86.71\%), showing that it remains competitive across both ID accuracy and OOD AUROC within the evaluated ResNet-18 setting. Although AP (86.71\%) and Prototype (86.07\%) remain strong in ID accuracy, neither surpasses Cross-Entropy for OOD detection. Triplet continues to struggle, with 68.12\% ID accuracy and 74.44\%/84.70\% near-/far-OOD AUROC, likely because forming informative triplets becomes harder in a large and diverse class set, leading to comparatively weaker embeddings.

\FloatBarrier

\subsection{Qualitative Analysis}
In this section, we provide t-SNE \cite{van2008visualizing} visualizations of embeddings for ID versus near- and far-OOD comparisons generated by models trained with different loss functions.
Appendix~\ref{app:extended_qual} provides extended qualitative analysis, including UMAP \cite{mcinnes2018umap-software} embeddings for ID versus OOD, and ID versus near- and far-OOD comparisons.

Both t-SNE and UMAP are effective for visualizing high-dimensional data, but they rely on different principles and have distinct advantages. t-SNE is particularly adept at revealing local structures but may distort global relationships, whereas UMAP provides a more balanced view of both local and global structures while improving computational efficiency. 
Because t-SNE and UMAP do not preserve exact pairwise distances, we use them only qualitatively to compare how training objectives shape the clustering and separation of in-distribution (ID) and out-of-distribution (OOD) samples. Visual patterns, such as tighter ID clustering or better OOD separation, can reveal valuable insights into the discriminative capacity and generalization behavior of learned embeddings, thus supporting and enriching our quantitative results.

For better observability, we use CIFAR-10 as the ID dataset. CIFAR-100 and TinyImageNet are used as near-OOD datasets, while MNIST, SVHN, Textures, and Places365 are used as far-OOD datasets. Since OOD samples become harder to detect as they become more similar to the training data, we analyze the near- and far-OOD cases separately.

\begin{figure}[!t]
    \centering
    \hfill
    \begin{subfigure}[t]{0.355\linewidth}
        \centering
        \includegraphics[width=\linewidth]{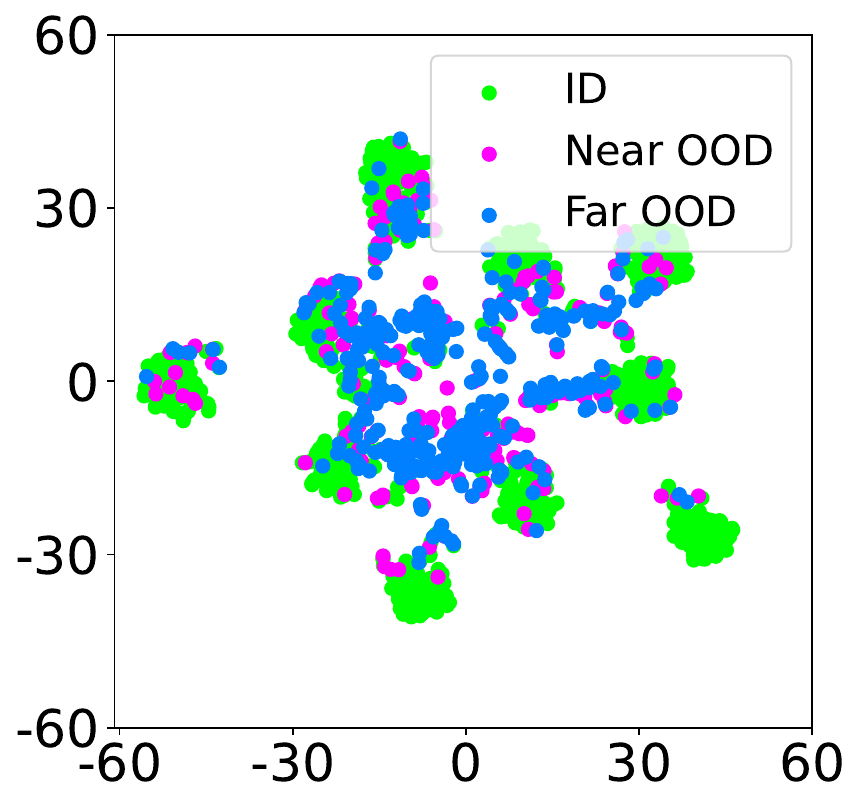}
        \caption{Cross-Entropy Loss}
    \end{subfigure}
    \hfill
    \begin{subfigure}[t]{0.34\linewidth}
        \centering
        \includegraphics[width=\linewidth]{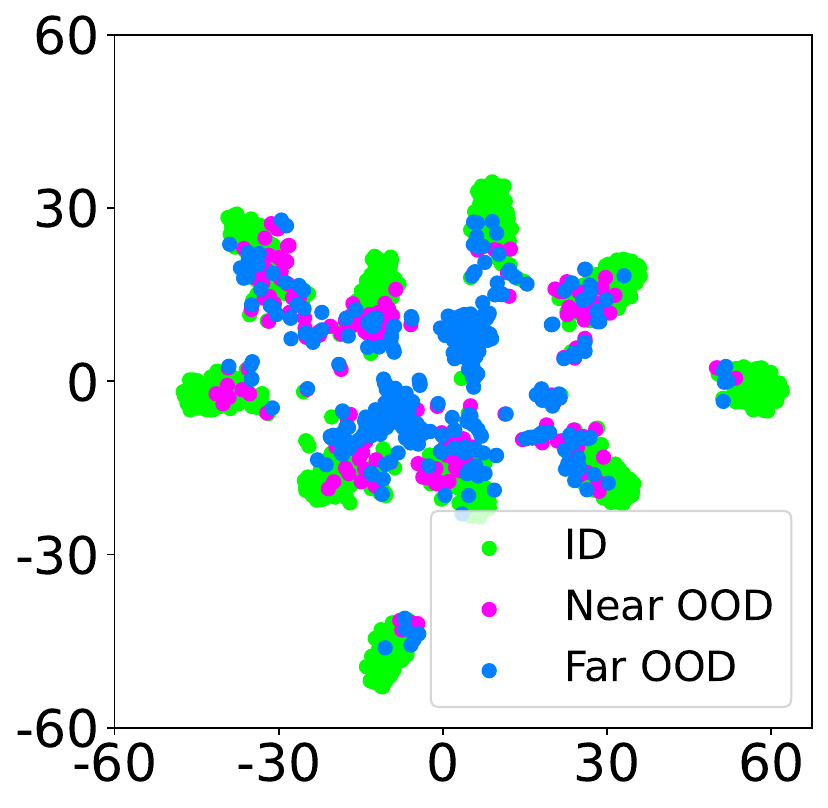}
        \caption{Triplet Loss}
    \end{subfigure}
    \hfill
    \vspace{1.2em}

    \hspace{3.2em}
    \begin{subfigure}[t]{0.34\linewidth}
        \centering
        \includegraphics[width=\linewidth]{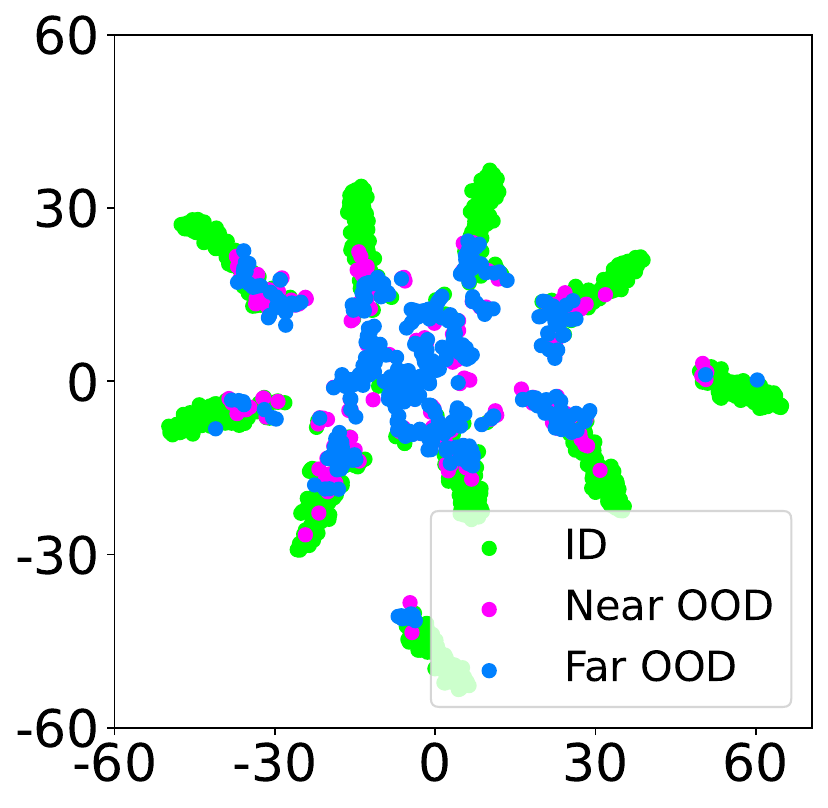}
        \caption{Prototype Loss}
    \end{subfigure}
    \hspace{3.8em}
    \begin{subfigure}[t]{0.353\linewidth}
        \centering
        \includegraphics[width=\linewidth]{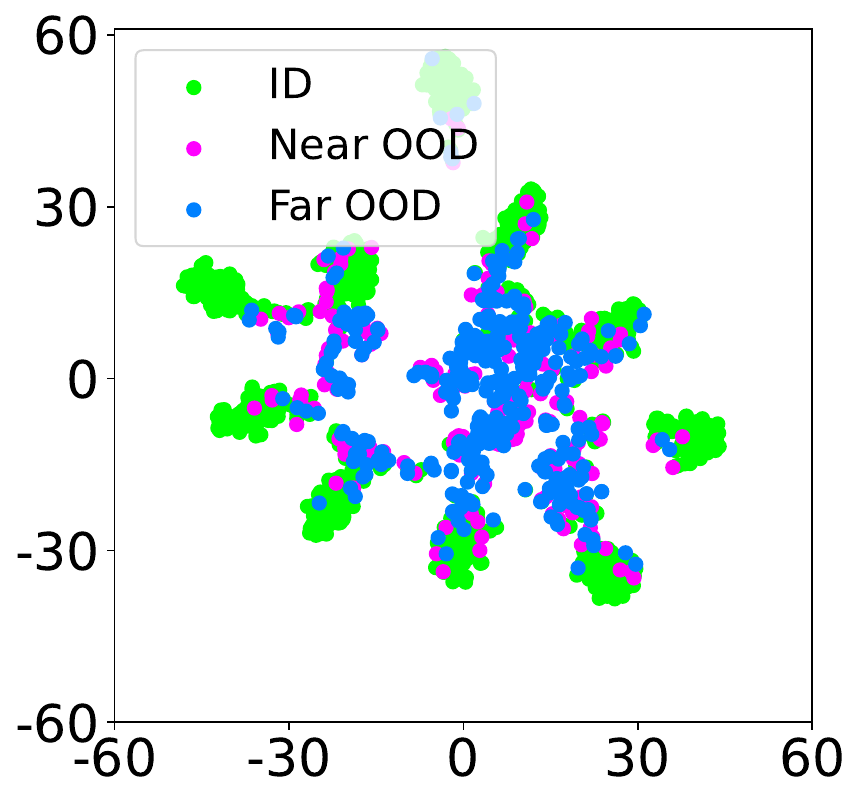}
        \caption{AP Loss}
    \end{subfigure}
    \hfill
    \vspace{-0.1em}
    \caption{t-SNE visualizations for ID vs near-OOD and far-OOD samples.}
    \label{fig:tsne_id_vs_near_and_far_ood}
\end{figure}

Figure~\ref{fig:tsne_id_vs_near_and_far_ood} shows ID, near-OOD, and far-OOD samples in the projected embedding spaces. Far-OOD instances tend to occupy regions visually separated from dense ID clusters and are not strongly aligned with any single ID class, consistent with their substantially different visual features. While these projected relationships are only qualitative, they provide an intuitive visualization that complements the OOD scores computed in the original model output and feature spaces:

\paragraph{Maximum Softmax Probability (MSP):} Far-OOD samples typically do not strongly match any ID class and therefore tend to receive lower maximum softmax probabilities than ID samples. Consequently, their negated-MSP OOD scores are higher, making them easier to distinguish from ID samples.

\paragraph{Minimum Distance to Prototypes/Train Samples:} Far-OOD samples tend to have larger distances to ID prototypes/training samples in the learned embedding space than ID instances. Hence, distance to the nearest ID prototype/sample can be used as an OOD score to distinguish far-OOD from ID samples.
    
\paragraph{Entropy of Softmax over Distances/Probabilities:} Far-OOD samples often do not align strongly with a single ID class and may have similar distances or probabilities across multiple classes. This leads to higher entropy, since entropy increases as the predicted probabilities become closer to each other.

Near-OOD samples, in contrast, tend to appear closer to the ID clusters, consistent with their greater visual similarity to the training data. This makes them harder to distinguish from ID samples and is consistent with the quantitative results, where far-OOD AUROC is typically higher than near-OOD AUROC. Overall, the qualitative analysis helps interpret the ID/OOD structure, but it does not clearly indicate which objective should dominate the others.

\section{Discussion}
The results reveal trade-offs across the evaluated objective-inference pairings in ID accuracy and OOD detection.

\begin{itemize}
    \item \textbf{Cross-Entropy Loss:} Cross-Entropy remains a strong baseline across all three datasets, with consistently high ID accuracy and OOD AUROC.
    \item \textbf{Triplet Loss:} Triplet is competitive in some OOD settings but shows substantially lower ID accuracy and near-OOD AUROC on CIFAR-100 and ImageNet-200.
    \item \textbf{Prototype Loss:} Prototype achieves strong ID accuracy, particularly on CIFAR-10 and CIFAR-100, but does not consistently improve OOD AUROC over Cross-Entropy.
    \item \textbf{Average Precision (AP) Loss:} AP achieves strong ID accuracy and competitive AUROC in several settings, but does not surpass Cross-Entropy for OOD detection on ImageNet-200.
\end{itemize}

\paragraph{Relation to prior findings.} Our distance-based inference for Triplet training aligns with the motivation of deep nearest-neighbor OOD scoring~\cite{sun2022ood_knn}, but our results suggest that triplet-based training becomes harder to scale as class count grows, which can offset the benefits of metric structure. Unlike works that study pre-training objectives such as masked image modeling \cite{li2023rethinking}, we keep the backbone and training pipeline fixed to improve comparability across the evaluated supervised objective-inference pairings. Compared to OOD-tailored objectives that reshape embedding geometry, such as hyperspherical embeddings~\cite{ming2023hyperspherical} and mixtures of prototypes~\cite{lu2024mixture}, our findings provide baseline trade-offs for common supervised objectives under the same OpenOOD protocols.

\paragraph{Computational Considerations.} Triplet additionally requires triplet mining and nearest-neighbor inference over training embeddings. Since we do not benchmark training time, inference time, or memory use, we draw no quantitative conclusions about computational efficiency.

\paragraph{Limitations.} Our study uses a single ResNet-18 backbone and a modest hyperparameter search, so the observed trends may not generalize to other architectures or broader optimization settings. Moreover, the training objective and OOD scoring rule are not fully factorized, and the reported OOD results therefore characterize objective-inference pairings rather than the training objective alone. Finally, our evaluation focuses on semantic near-/far-OOD shifts and AUROC; covariate-shift settings, stronger post-hoc scoring methods, operating-point and precision-recall metrics such as FPR95 and AUPR, and quantitative analyses of representation geometry remain outside the scope of this study.

\section{Conclusion}
In this study, we presented a systematic comparison of four widely used training objectives: Cross-Entropy Loss, Triplet Loss, Prototype Loss, and Average Precision (AP) Loss, representing probabilistic classification, margin-based metric learning, prototype-based learning, and ranking-based optimization. Under a fixed ResNet-18 architecture and standardized OpenOOD evaluation, the evaluated objective-inference pairings show that Cross-Entropy provides the most consistent overall balance between ID accuracy and near- and far-OOD AUROC, while the alternative objectives remain competitive in specific settings.

Because the objectives are evaluated with scoring rules aligned with their respective output spaces, our results characterize the practical behavior of these objective-inference pairings rather than fully isolating the effect of the training objective alone. Accordingly, our conclusions are limited to the evaluated backbone, OOD settings, and AUROC-based comparison. Within this scope, the results clarify the trade-offs among the four supervision paradigms and support Cross-Entropy as a strong default for the OpenOOD setting studied.

\bibliographystyle{splncs04}
\bibliography{references}

\clearpage
\appendix
\raggedbottom
\setlength{\intextsep}{12pt plus 2pt minus 2pt}
\captionsetup[table]{skip=4pt}
\section{Selected Hyperparameters and OOD Post-Processing}
\label{app:selected_params}

For each dataset-objective pair, model hyperparameters are selected using classification accuracy on the predefined OpenOOD ID validation split. For Cross-Entropy and Prototype, MSP versus predictive entropy is selected using AUROC on the dedicated OpenOOD ID- and OOD-validation splits. The OOD-validation set is TinyImageNet for CIFAR-10/100 and OpenImage-O for ImageNet-200. These validation splits are separate from the near- and far-OOD test splits, which are used only for final reporting. We tune the learning rate (\textbf{LR}), embedding dimension (\textbf{ED}) for embedding-based objectives, and the Prototype parameters \(\boldsymbol{\lambda}\) and \(\boldsymbol{\tau}\).

Tables~\ref{tab:selected_params_cifar10}--\ref{tab:selected_params_imagenet200} report the selected settings. Cross-Entropy and Prototype use the selected MSP/Entropy rule, Triplet uses 1-NN distance to the training embeddings, and AP uses softmax entropy. Triplet mining is Random for CIFAR-10 and Semi-Hard for CIFAR-100 and ImageNet-200.

\begin{table}[H]
\centering
\begin{tabular*}{\textwidth}{@{\extracolsep{\fill}}lccccc@{}}
\toprule
\textbf{Objective} & \textbf{LR} & \textbf{ED} & \(\boldsymbol{\lambda}\) / \(\boldsymbol{\tau}\) & \textbf{Triplet Mining} & \textbf{OOD score} \\
\midrule
AP Loss             & 0.08  & --  & -- / --        & --        & Entropy \\
Cross-Entropy Loss  & 0.10  & --  & -- / --        & --        & Entropy \\
Prototype Loss      & 0.10  & 64  & 0.01 / 0.1     & --        & Entropy \\
Triplet Loss        & 0.005 & 32  & -- / --        & Random    & 1-NN distance \\
\bottomrule
\end{tabular*}
\caption{Selected hyperparameters and OOD post-processing for CIFAR-10.}
\label{tab:selected_params_cifar10}
\end{table}

\begin{table}[H]
\centering
\begin{tabular*}{\textwidth}{@{\extracolsep{\fill}}lccccc@{}}
\toprule
\textbf{Objective} & \textbf{LR} & \textbf{ED} & \(\boldsymbol{\lambda}\) / \(\boldsymbol{\tau}\) & \textbf{Triplet Mining} & \textbf{OOD score} \\
\midrule
AP Loss             & 0.08   & --  & -- / --        & --        & Entropy \\
Cross-Entropy Loss  & 0.08   & --  & -- / --        & --        & Entropy \\
Prototype Loss      & 0.10   & 128 & 0.001 / 0.1    & --        & MSP \\
Triplet Loss        & 0.0004 & 256 & -- / --        & Semi-Hard & 1-NN distance \\
\bottomrule
\end{tabular*}
\caption{Selected hyperparameters and OOD post-processing for CIFAR-100.}
\label{tab:selected_params_cifar100}
\end{table}

\begin{table}[H]
\centering
\begin{tabular*}{\textwidth}{@{\extracolsep{\fill}}lccccc@{}}
\toprule
\textbf{Objective} & \textbf{LR} & \textbf{ED} & \(\boldsymbol{\lambda}\) / \(\boldsymbol{\tau}\) & \textbf{Triplet Mining} & \textbf{OOD score} \\
\midrule
AP Loss             & 0.10   & --  & -- / --        & --        & Entropy \\
Cross-Entropy Loss  & 0.08   & --  & -- / --        & --        & Entropy \\
Prototype Loss      & 0.10   & 128 & 0.001 / 0.1    & --        & MSP \\
Triplet Loss        & 0.0009 & 512 & -- / --        & Semi-Hard & 1-NN distance \\
\bottomrule
\end{tabular*}
\caption{Selected hyperparameters and OOD post-processing for ImageNet-200.}
\label{tab:selected_params_imagenet200}
\end{table}

\clearpage
\section{Extended Quantitative Results}
\label{app:extended_quant}

In this section, we report extended results (mean \(\pm\) standard deviation) for each training objective across five independent runs on the CIFAR-10, CIFAR-100, and ImageNet-200 OpenOOD benchmarks (Tables~\ref{table:cifar10_openood}--\ref{table:imagenet200_openood}). The best performance in each column is shown in \textbf{bold}, while the second-best is \underline{underlined}. The corresponding selected hyperparameters and OOD post-processing rules for these runs are provided in Appendix~\ref{app:selected_params} (Tables~\ref{tab:selected_params_cifar10}--\ref{tab:selected_params_imagenet200}).

\begin{table}[H]
\centering
\begin{tabular*}{\textwidth}{@{\extracolsep{\fill}}lccc@{}}
\toprule
\textbf{Loss Function} & \textbf{ID Accuracy (\%)} & \textbf{Near-OOD AUROC (\%)} & \textbf{Far-OOD AUROC (\%)} \\
\midrule
AP Loss           & 94.93\(\pm\)0.20            & \underline{88.80\(\pm\)0.19} & \textbf{92.34\(\pm\)0.53} \\
Cross-Entropy Loss & \underline{94.95\(\pm\)0.25} & 88.23\(\pm\)0.42           & 90.98\(\pm\)0.47 \\
Prototype Loss    & \textbf{95.06\(\pm\)0.14}    & 87.87\(\pm\)0.37           & 91.28\(\pm\)0.51 \\
Triplet Loss      & 93.56\(\pm\)0.18            & \textbf{88.85\(\pm\)0.20}  & \underline{92.25\(\pm\)0.58} \\
\bottomrule
\end{tabular*}
\caption{Extended performance comparison on CIFAR-10 OpenOOD benchmark.}
\label{table:cifar10_openood}
\end{table}

\begin{table}[H]
\centering
\begin{tabular*}{\textwidth}{@{\extracolsep{\fill}}lccc@{}}
\toprule
\textbf{Loss Function} & \textbf{ID Accuracy (\%)} & \textbf{Near-OOD AUROC (\%)} & \textbf{Far-OOD AUROC (\%)} \\
\midrule
AP Loss           & \underline{77.21\(\pm\)0.46} & 79.32\(\pm\)0.27           & 77.68\(\pm\)1.15 \\
Cross-Entropy Loss & 77.10\(\pm\)0.28           & \textbf{80.94\(\pm\)0.24}  & \underline{80.46\(\pm\)0.75} \\
Prototype Loss    & \textbf{78.18\(\pm\)0.54}    & \underline{79.95\(\pm\)0.33} & 79.85\(\pm\)1.42 \\
Triplet Loss      & 69.00\(\pm\)0.30            & 76.95\(\pm\)0.27           & \textbf{80.50\(\pm\)1.05} \\
\bottomrule
\end{tabular*}
\caption{Extended performance comparison on CIFAR-100 OpenOOD benchmark.}
\label{table:cifar100_openood}
\end{table}

\begin{table}[H]
\centering
\begin{tabular*}{\textwidth}{@{\extracolsep{\fill}}lccc@{}}
\toprule
\textbf{Loss Function} & \textbf{ID Accuracy (\%)} & \textbf{Near-OOD AUROC (\%)} & \textbf{Far-OOD AUROC (\%)} \\
\midrule
AP Loss           & \textbf{86.71\(\pm\)0.30}    & \underline{82.47\(\pm\)0.20} & \underline{90.54\(\pm\)0.28} \\
Cross-Entropy Loss & \underline{86.66\(\pm\)0.22} & \textbf{83.54\(\pm\)0.19}   & \textbf{91.16\(\pm\)0.25} \\
Prototype Loss    & 86.07\(\pm\)0.34            & 80.78\(\pm\)0.17           & 89.73\(\pm\)0.23 \\
Triplet Loss      & 68.12\(\pm\)0.22            & 74.44\(\pm\)0.14           & 84.70\(\pm\)0.23 \\
\bottomrule
\end{tabular*}
\caption{Extended performance comparison on ImageNet-200 OpenOOD benchmark.}
\label{table:imagenet200_openood}
\end{table}

\clearpage
\section{Extended Qualitative Analysis}
\label{app:extended_qual}

\subsection{ID vs.\ OOD Comparison}
To complement the quantitative results, we analyze the embedding geometry induced by different training objectives using t-SNE and UMAP projections.

\begin{figure}[!t]
    \centering
    \hfill
    \begin{subfigure}[t]{0.355\linewidth}
        \centering
        \includegraphics[width=\linewidth]{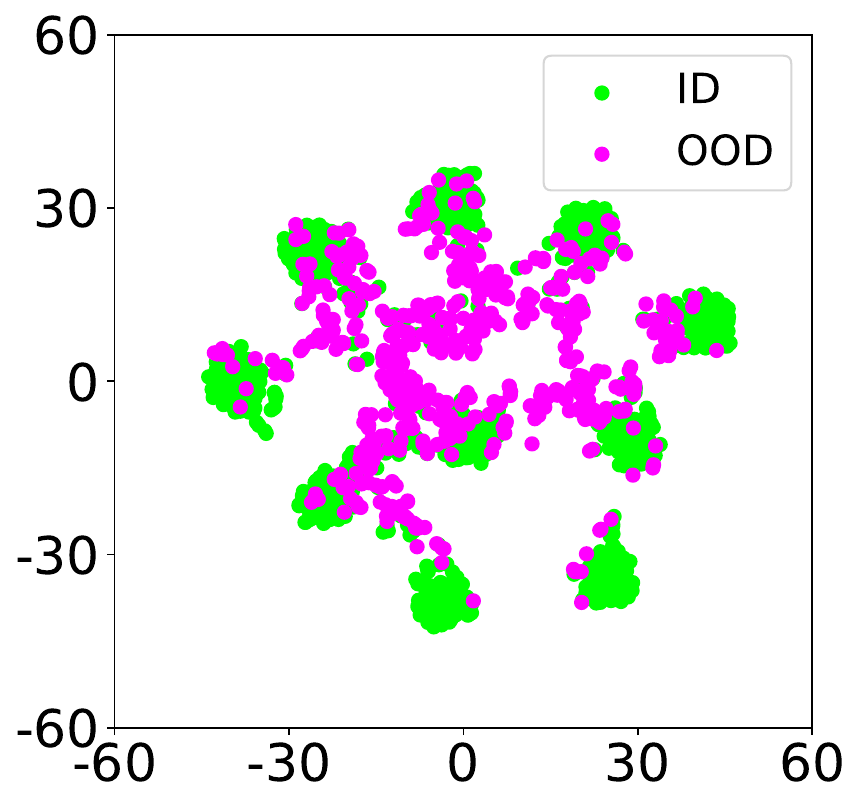}
        \caption{Cross-Entropy Loss}
    \end{subfigure}
    \hfill
    \begin{subfigure}[t]{0.34\linewidth}
        \centering
        \includegraphics[width=\linewidth]{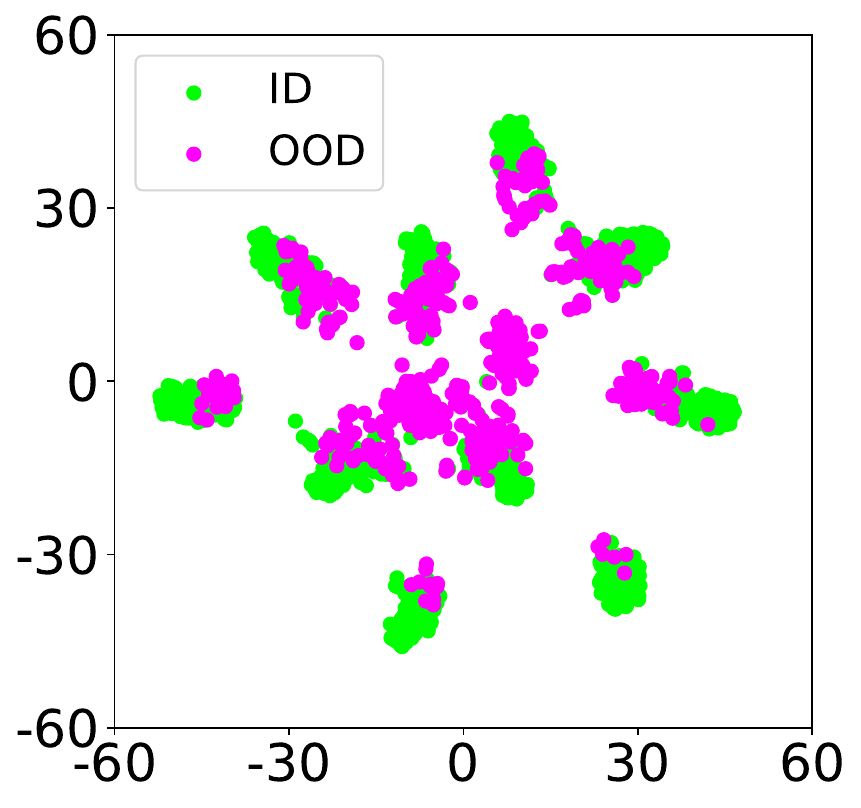}
        \caption{Triplet Loss}
    \end{subfigure}
    \hfill
    \vspace{1.2em}

    \hspace{3.2em}
    \begin{subfigure}[t]{0.34\linewidth}
        \centering
        \includegraphics[width=\linewidth]{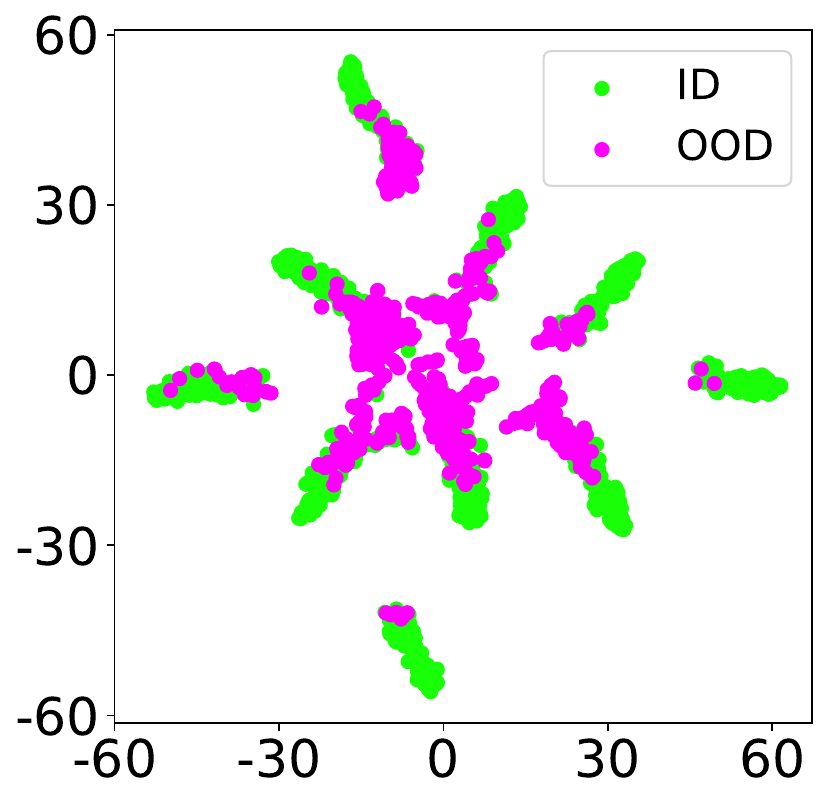}
        \caption{Prototype Loss}
    \end{subfigure}
    \hspace{3.8em}
    \begin{subfigure}[t]{0.353\linewidth}
        \centering
        \includegraphics[width=\linewidth]{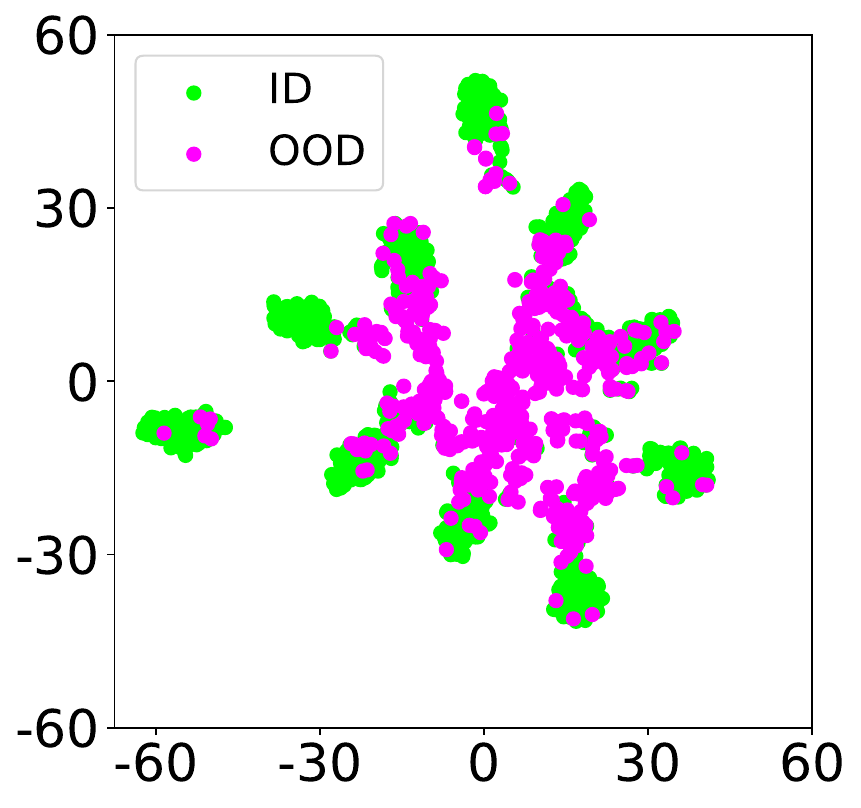}
        \caption{AP Loss}
    \end{subfigure}
    \hfill
    \vspace{-0.1em}
    \caption{t-SNE visualizations for ID vs OOD comparison.}
    \label{fig:tsne_id_vs_ood}
\end{figure}

Figure~\ref{fig:tsne_id_vs_ood} illustrates the distribution of ID and OOD samples produced by models trained with different loss functions. The projections are generated using t-SNE, with colors indicating ID and OOD samples.

In the two-dimensional projections, the in-distribution classes appear as relatively compact clusters, with samples visually grouped together and separated from other clusters. While dimensionality reduction methods do not preserve exact geometric relationships, this qualitative pattern is broadly consistent with the high ID accuracy observed on CIFAR-10 (all methods exceeding 93\%), suggesting that the learned representations remain discriminative for in-distribution classification.

In the projections, the ID clusters appear relatively evenly distributed across the two-dimensional space, which makes it difficult to infer detailed semantic relationships between classes from the visualization alone.

In the projected embedding space, many OOD samples appear separated from the main ID clusters, reflecting their distributional difference from the training data. However, some OOD instances lie close to specific ID clusters, indicating that some OOD datasets contain samples with visual features similar to those of the in-distribution classes, which can make detection more difficult.

\begin{figure}[!t]
    \centering
    \hfill
    \begin{subfigure}[t]{0.355\linewidth}
        \centering
        \includegraphics[width=\linewidth]{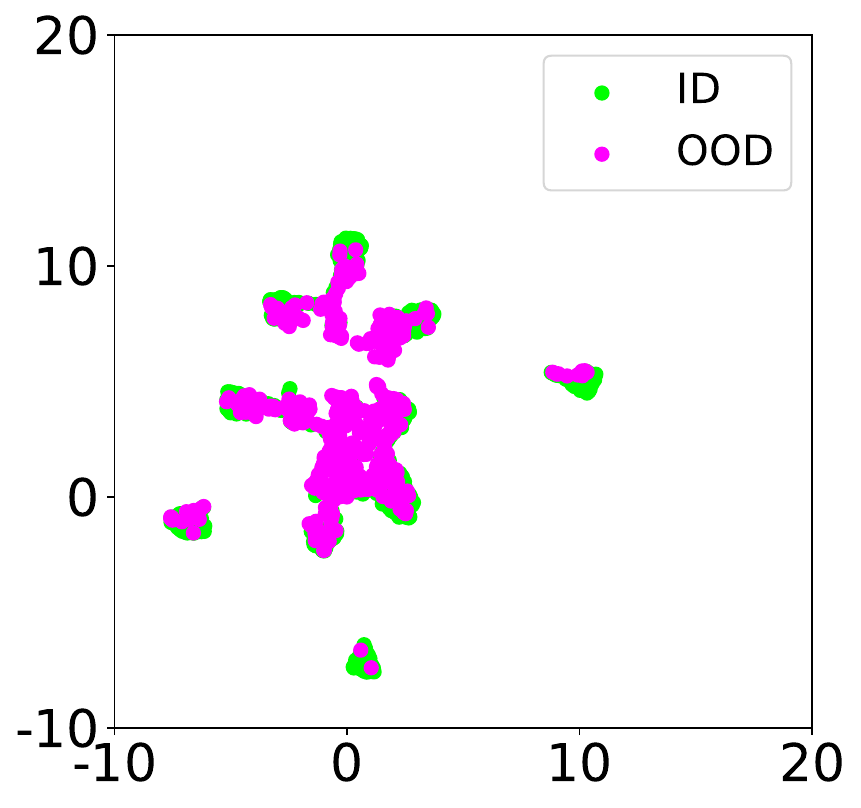}
        \caption{Cross-Entropy Loss}
    \end{subfigure}
    \hfill
    \begin{subfigure}[t]{0.34\linewidth}
        \centering
        \includegraphics[width=\linewidth]{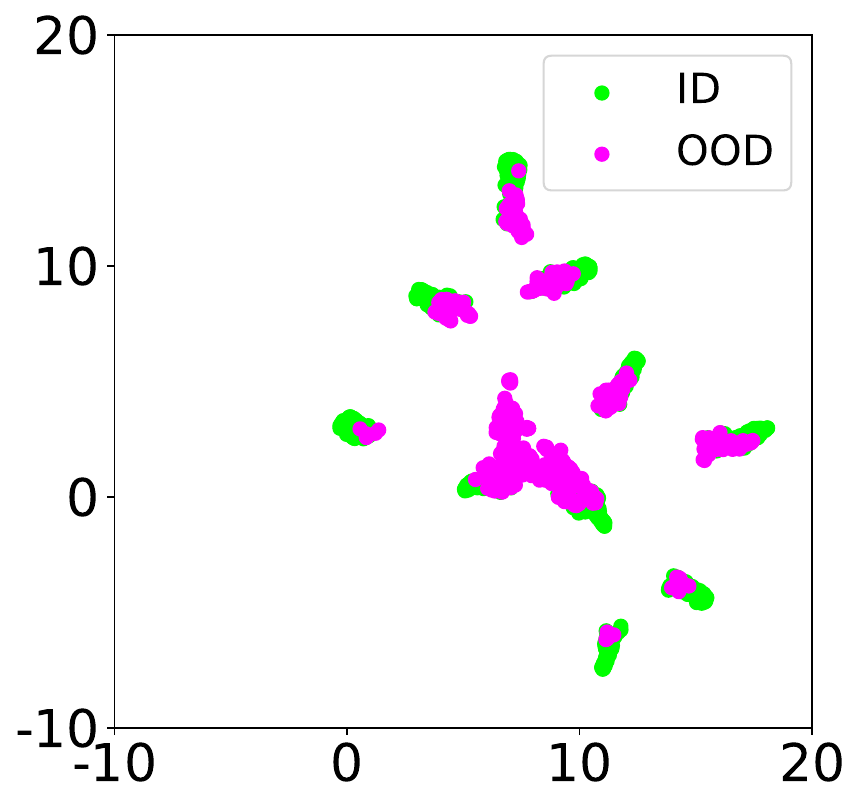}
        \caption{Triplet Loss}
    \end{subfigure}
    \hfill
    \vspace{1.2em}

    \hspace{3.2em}
    \begin{subfigure}[t]{0.34\linewidth}
        \centering
        \includegraphics[width=\linewidth]{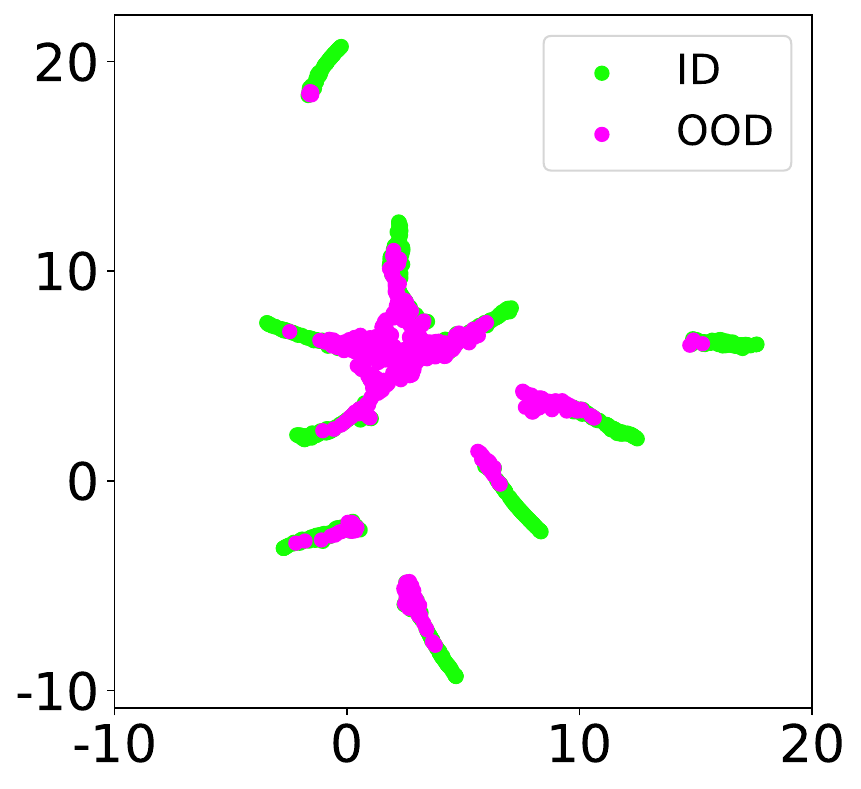}
        \caption{Prototype Loss}
    \end{subfigure}
    \hspace{3.8em}
    \begin{subfigure}[t]{0.353\linewidth}
        \centering
        \includegraphics[width=\linewidth]{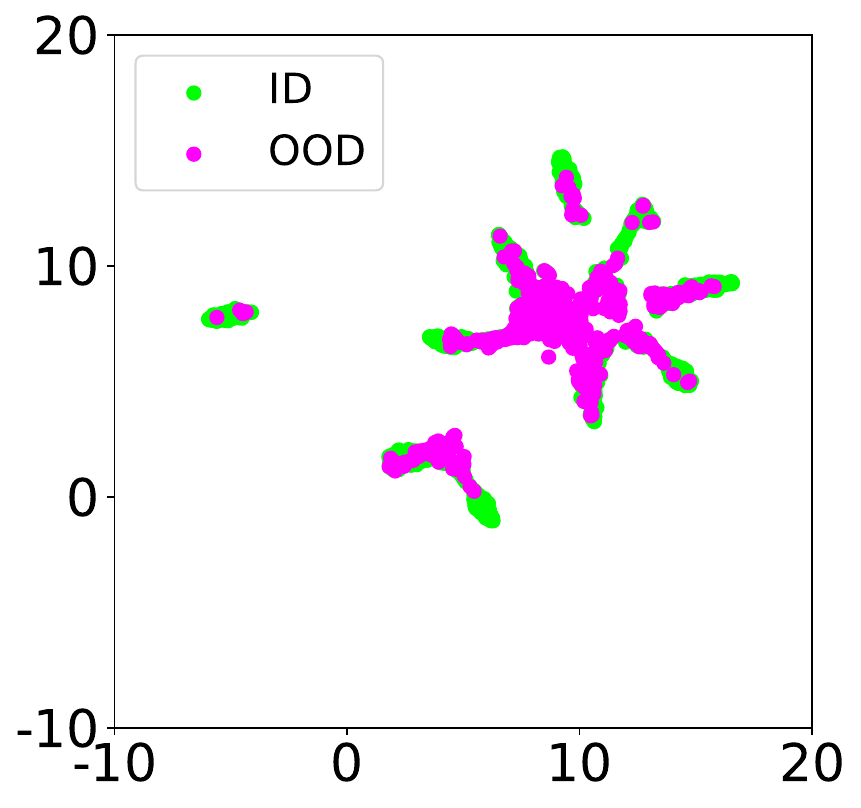}
        \caption{AP Loss}
    \end{subfigure}
    \hfill
    \vspace{-0.1em}
    \caption{UMAP visualizations for ID vs OOD comparison.}
    \label{fig:umap_id_vs_ood}
\end{figure}

Figure~\ref{fig:umap_id_vs_ood} shows UMAP projections of the embeddings, providing a complementary view of both local and broader structural patterns in the data.

In comparison to the t-SNE results, where the ID clusters are distributed relatively homogeneously, the UMAP visualizations provide additional insight into the relative arrangement of ID clusters in the two-dimensional space, offering qualitative cues about potential semantic relationships between classes in CIFAR-10.

Similar to the t-SNE projections, many OOD samples appear positioned away from the dense ID clusters and are not strongly aligned with any single ID class. However, the presence of OOD data points close to the ID clusters is more pronounced in the UMAP visualizations, suggesting that detecting some OOD samples may be challenging, as their embeddings are similar to those of the ID samples.

\subsection{ID vs.\ Near-OOD and Far-OOD Comparison}

\begin{figure}[H]
    \centering
    \hfill
    \begin{subfigure}[t]{0.355\linewidth}
        \centering
        \includegraphics[width=\linewidth]{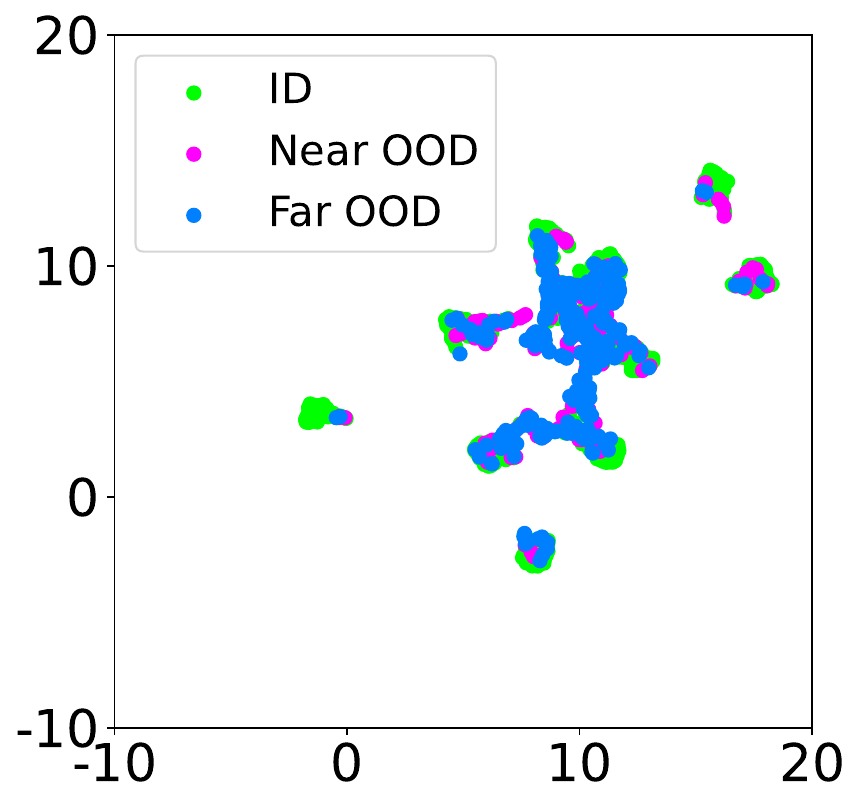}
        \caption{Cross-Entropy Loss}
    \end{subfigure}
    \hfill
    \begin{subfigure}[t]{0.34\linewidth}
        \centering
        \includegraphics[width=\linewidth]{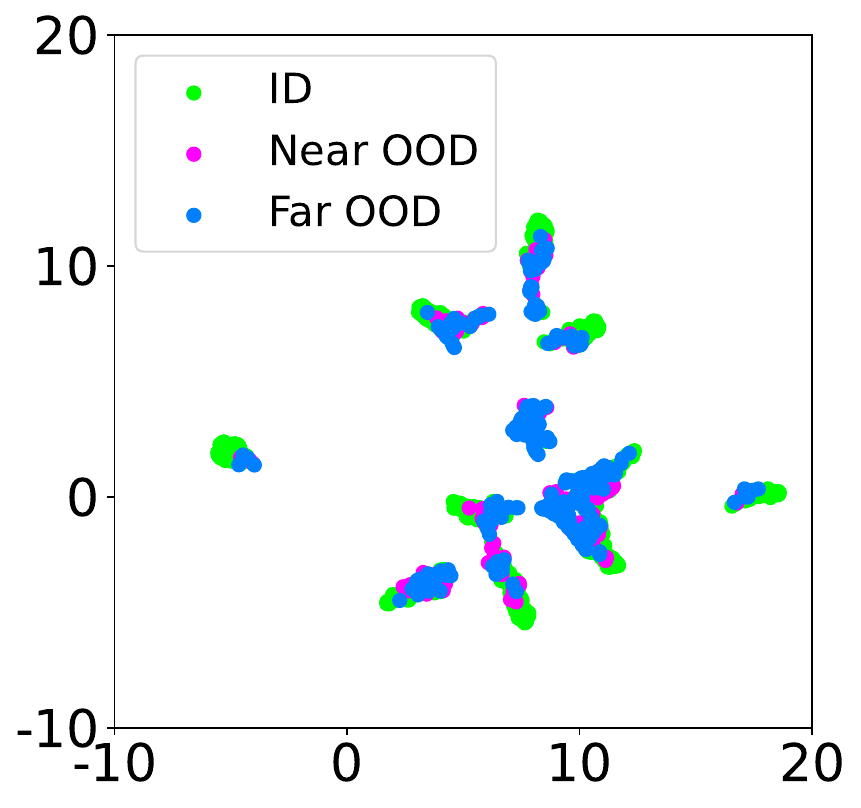}
        \caption{Triplet Loss}
    \end{subfigure}
    \hfill
    \vspace{1.2em}

    \hspace{3.2em}
    \begin{subfigure}[t]{0.34\linewidth}
        \centering
        \includegraphics[width=\linewidth]{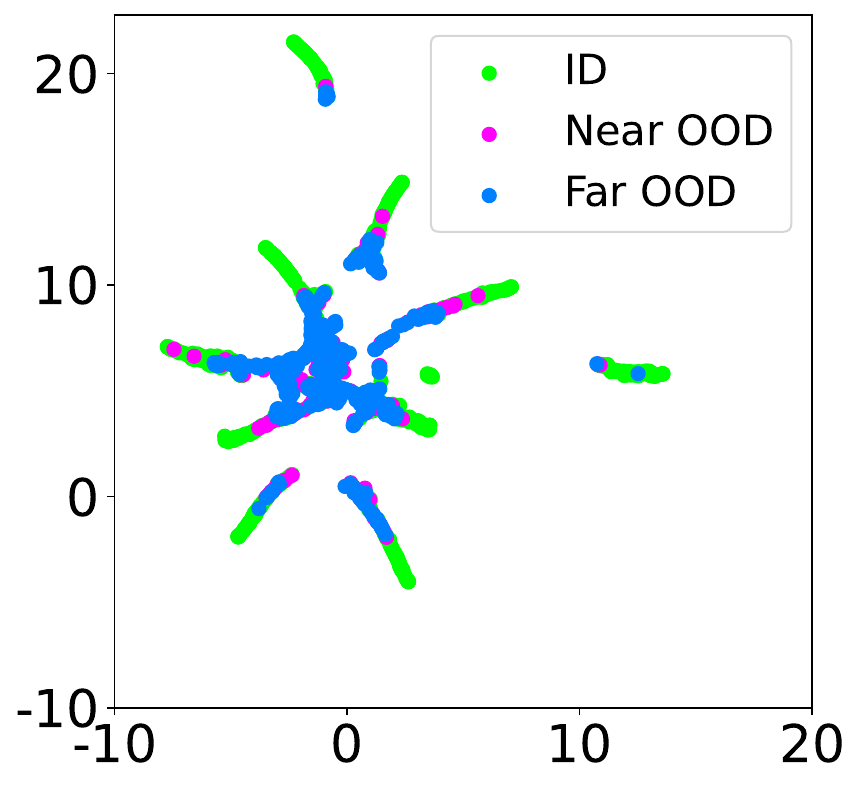}
        \caption{Prototype Loss}
    \end{subfigure}
    \hspace{3.8em}
    \begin{subfigure}[t]{0.353\linewidth}
        \centering
        \includegraphics[width=\linewidth]{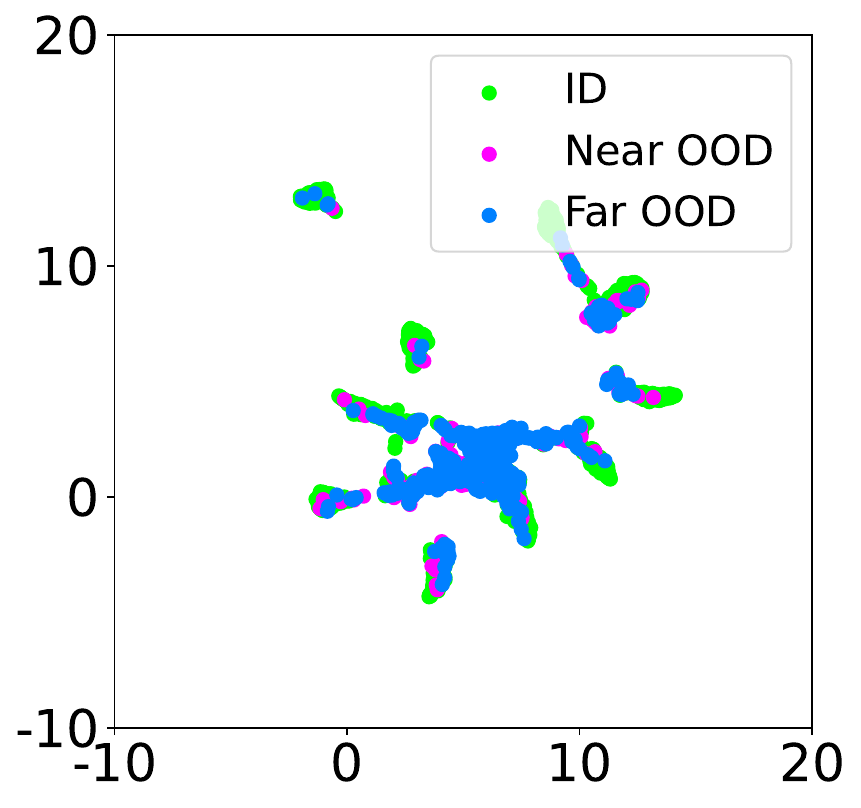}
        \caption{AP Loss}
    \end{subfigure}
    \hfill
    \vspace{-0.1em}
    \caption{UMAP visualizations for ID vs near-OOD and far-OOD comparison.}
    \label{fig:umap_id_vs_near_and_far_ood}
\end{figure}

Figure~\ref{fig:umap_id_vs_near_and_far_ood} shows UMAP embeddings for in-distribution (ID), near-OOD, and far-OOD samples. In the projections, far-OOD samples tend to appear more separated from the main ID clusters, whereas near-OOD samples often appear closer to certain ID clusters. These patterns are qualitatively consistent with our detection strategies and complement our quantitative findings.

\end{document}